\pdfoutput=1
\documentclass{article}

\usepackage{afterpage}
\usepackage{arxiv}

\usepackage[utf8]{inputenc} 
\usepackage[T1]{fontenc}    
\usepackage{hyperref}       
\usepackage{url}            
\usepackage{booktabs}       
\usepackage{amsfonts}       
\usepackage{nicefrac}       
\usepackage{microtype}      
\usepackage{lipsum}
\usepackage{graphicx}
\usepackage{times}
\usepackage{soul}
\usepackage{url}
\usepackage{graphicx}
\usepackage{amsthm}
\usepackage{dsfont}

\usepackage{amsmath}
\usepackage{amsthm}
\usepackage{booktabs}
\usepackage{algorithm}
\usepackage{algpseudocode}
\usepackage{algorithmicx}
\usepackage[switch]{lineno}
\usepackage{natbib}

\usepackage{mathtools}
\usepackage{amssymb}
\usepackage{xcolor}
\usepackage{multirow}
\usepackage{pifont}
\usepackage{threeparttable}
\graphicspath{{./figures/}}
\usepackage{subcaption}

\title{Uncertainty Quantification for Hallucination Detection in Large Language Models: Foundations, Methodology, and Future Directions}

\author{
Sungmin Kang \\
University of Southern California\\
\texttt{kangsung@usc.edu} 
\And
Yavuz Faruk Bakman \\
University of Southern California\\
\texttt{ybakman@usc.edu} 
\AND
Duygu Nur Yaldiz\\
University of Southern California\\
\texttt{yaldiz@usc.edu}
\And
Baturalp Buyukates \\
University of Birmingham\\
\texttt{b.buyukates@bham.ac.uk} \\
\And
Salman Avestimehr \\
University of Southern California\\
\texttt{avestime@usc.edu} \\
}

\begin{document}
\maketitle
\begin{abstract}
The rapid advancement of large language models (LLMs) has transformed the landscape of natural language processing, enabling breakthroughs across a wide range of areas including question answering, machine translation, and text summarization.
Yet, their deployment in real-world applications has raised concerns over reliability and trustworthiness, as LLMs remain prone to hallucinations that produce plausible but factually incorrect outputs.
Uncertainty quantification (UQ) has emerged as a central research direction to address this issue, offering principled measures for assessing the trustworthiness of model generations.
We begin by introducing the foundations of UQ, from its formal definition to the traditional distinction between epistemic and aleatoric uncertainty, and then highlight how these concepts have been adapted to the context of LLMs. 
Building on this, we examine the role of UQ in hallucination detection, where quantifying uncertainty provides a mechanism for identifying unreliable generations and improving reliability.
We systematically categorize a wide spectrum of existing methods along multiple dimensions and present empirical results for several representative approaches. 
Finally, we discuss current limitations and outline promising future research directions, providing a clearer picture of the current landscape of LLM UQ for hallucination detection.
\end{abstract}

\section{Introduction}

Generative large language models (LLMs) have gained significant attention for their remarkable ability to understand and generate human language with high accuracy and scalability. 
Given their effectiveness, these models have been widely adopted across various natural language processing tasks, including machine translation, text summarization, and question answering. 
Despite their success, LLMs are prone to producing erroneous or misleading outputs, often referred to as \textit{hallucinations}. 
This drawback poses a substantial limitation especially in high-stakes domains such as healthcare, law, and marketing~\cite{bengio2025internationalaisafetyreport, ravi2024lynxopensourcehallucination}, where reliable question answering is critical.
To ensure trustworthiness of LLMs, researchers have investigated various approaches, focusing on understanding hallucination mechanisms, detecting hallucinations in generated outputs, and mitigating them pre-generation. 
Among these, hallucination detection has emerged as a particularly important line of work, as it directly addresses the reliability of model outputs in real-world applications.

Hallucination detection refers to the task of identifying whether the outputs generated by LLMs are faithful to the input or contain fabricated or unsupported information. 
Approaches such as fact verification~\cite{shuster-etal-2021-retrieval-augmentation, chern2023factool, jiang-etal-2023-active}, \textit{LLM-as-a-judge}~\cite{10.5555/3666122.3668142, cohen-etal-2023-lm, feng-etal-2024-dont}, chain-of-thought~\cite{zhao-etal-2023-verify, dhuliawala-etal-2024-chain}, external tool-assisted verification~\cite{heyueya2023solvingmathwordproblems, chen-etal-2024-good}, and uncertainty quantification (UQ)~\cite{malinin2021uncertainty} are widely used for hallucination detection in generative LLMs.
Unlike other detection strategies that often depend on external resources such as the internet or secondary models for cross-checking, UQ inherently evaluates the reliability of generated content by internally assessing the confidence levels of the model’s responses, offering a significant advantage over other hallucination detection methods.
Thus, UQ is not limited to a binary decision of whether an output is hallucinated or not. 
Moreover, it quantifies the degree of certainty associated with each response, providing a finer-grained signal of trustworthiness.

UQ has been deeply studied in machine learning, particularly in classification scenarios. 
These traditional methods which rely on class probabilities, cannot be directly applied to LLMs because of their auto-regressive generative structure and open-ended output space, which pose fundamentally different challenges for defining and evaluating uncertainty. 
Nevertheless, the classical decomposition of uncertainty into \textit{aleatoric} and \textit{epistemic} forms remains a useful framework~\cite{H_llermeier_2021}.
Aleatoric (or data) uncertainty inherently arises from the variability of the data distribution, which is often considered irreducible. 
Epistemic (or model) uncertainty stems from a lack of knowledge or insufficient model capacity, and can in principle be reduced with more data or better modeling. 
Crucially, epistemic uncertainty is closely tied to hallucination in LLMs: when the model is forced to generate outputs in areas where it lacks sufficient knowledge, it is more likely to produce unsupported or fabricated content. 
Thus, quantifying epistemic uncertainty provides an effective means of assessing the trustworthiness of model outputs and serves as a valuable tool for hallucination detection.

In this study, we provide a comprehensive explanation of UQ as a tool for LLM hallucination detection on open-ended question-answering (QA) tasks. 
We seek to investigate the effectiveness of UQ by defining uncertainty for LLMs, systematically categorizing and comparing various approaches.
We provide a detailed explanation of how UQ methods are evaluated and present experimental results for several representative approaches.
Furthermore, we extend our analysis from short-form QA to long-form QA, which holds greater significance in practical applications yet remains under-explored.
Lastly, we devote a section for the limitations and future directions of UQ to offer a clearer understanding of its current landscape in hallucination detection.

Previous survey works have addressed UQ for generative LLMs from a broad perspective~\cite{huang2024trustllmtrustworthinesslargelanguage, huang2024surveyuncertaintyestimationllms}.
In contrast, our study focuses on the application of UQ as a tool for hallucination detection, enabling a more in-depth examination of UQ methodologies tailored to this specific challenge.
We systematically categorize existing works with an emphasis on UQ methods for QA tasks, while excluding studies on interactive agents or adaptive hallucination mitigation, despite their significant research value.
Additionally, our work presents a comprehensive experimental study that rigorously evaluates UQ methods on widely recognized QA datasets and across diverse language models.

\begin{figure*}[t]
\begin{center}
\includegraphics[width=\textwidth]{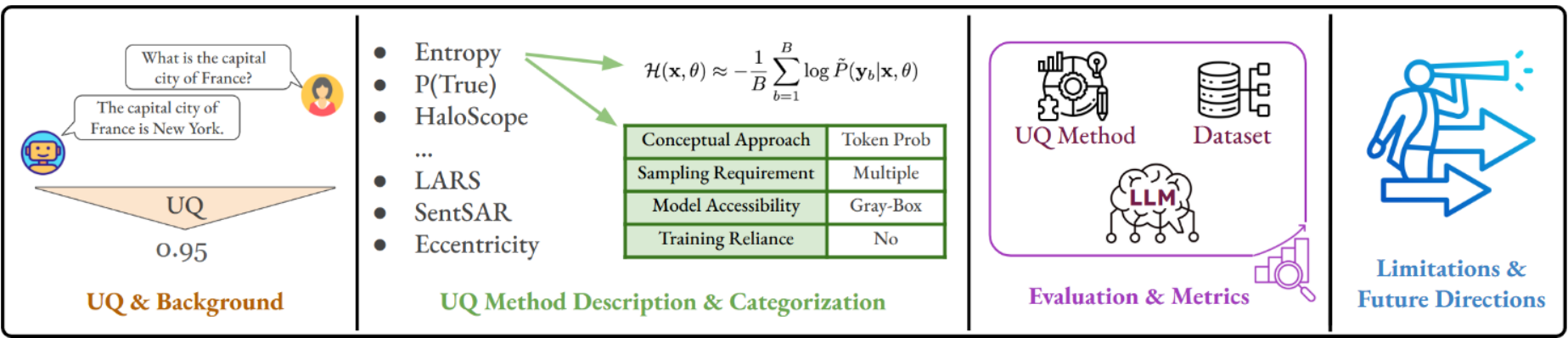}
\caption{Overview of this survey. We discuss 
the background of UQ, categorization and detailed description of existing methods, evaluation, limitations and future directions.}
\label{fig:overview}
\end{center}
\end{figure*}
\section{Background}
\label{sec:Background}

Uncertainty has been extensively studied in the field of deep learning, particularly in the context of classification. From a Bayesian perspective, UQ provides insights into model confidence and has been leveraged to improve decision-making in high-stakes applications. More recently, the scope of uncertainty research has expanded to LLMs, where estimating the uncertainty of generated answers has become a crucial problem. However, due to the auto-regressive generative structure of LLMs, traditional methods for UQ cannot be directly applied, necessitating new approaches designed for generative models.

In this section, we first introduce the concept of uncertainty and its role in machine learning. We then review UQ for classification tasks, followed by approaches developed for generative LLMs. Next, we describe the hallucination detection problem, which represents a key challenge in modern LLMs, and finally discuss how UQ has been adapted to address hallucination detection.

\subsection{Uncertainty in Machine Learning}

Uncertainty is a crucial concept in machine learning and artificial intelligence, but it remains inherently ambiguous and difficult to define precisely. Broadly, uncertainty refers to the degree of doubt or lack of confidence in a model’s prediction, reflecting situations where models may be overconfident in incorrect outputs or insufficiently calibrated to the underlying data distribution. Understanding and quantifying uncertainty is crucial for building trustworthy AI systems, especially in high-stakes domains such as medical diagnosis, autonomous driving, and natural language generation.

Modern deep learning models often achieve high prediction accuracy but can fail in complex or out-of-distribution scenarios, producing confident yet incorrect results. Such failures highlight the importance of knowing what the model does not know. UQ allows systems to detect when predictions may be unreliable, enabling such as calibration, human intervention, additional verification, or cautious decision-making. In generative systems like LLMs, uncertainty further plays a key role in assessing the factuality of outputs and mitigating hallucinations.

Uncertainty in machine learning is typically decomposed into two categories, aleatoric uncertainty and epistemic uncertainty.

\textbf{Aleatoric uncertainty} (also called \emph{data uncertainty}) arises from inherent randomness or noise in the data. It arises from the stochastic nature of the underlying data-generating process, such as measurement noise in sensors, ambiguous or conflicting labels provided by annotators, or overlapping class boundaries that cannot be perfectly separated. This form of uncertainty is considered irreducible~\cite{pmlr-v37-wilson15}, since the intrinsic randomness of the data cannot be fully eliminated even with unlimited training data. In natural language tasks, aleatoric uncertainty often manifests in situations where multiple responses may be equally valid or when the meaning of a phrase is inherently ambiguous, making it impossible for a model to assign certainty to a single deterministic outcome.

\textbf{Epistemic uncertainty} (also called \emph{model uncertainty}) reflects a model’s lack of knowledge about the task or domain. This type of uncertainty stems from limited or biased training data, suboptimal model architectures, or mismatches between training and test distributions (distributional shift). Unlike aleatoric uncertainty, epistemic uncertainty is considered reducible: it can be mitigated by collecting more representative training data, designing more expressive models, or incorporating prior knowledge. For instance, a model trained primarily on English may display high epistemic uncertainty when applied to a low-resource language, but this can be reduced by incorporating more multilingual data.

In practice, the total predictive uncertainty of a model can be viewed as the combination of aleatoric and epistemic components. Distinguishing between them is critical as aleatoric uncertainty highlights the intrinsic variability of the data, whereas epistemic uncertainty identifies gaps in the model’s knowledge. Together, these notions form the foundation for UQ methods, which will be further elaborated in the following discussion.

\begin{figure}[!htbp]
\centering
\includegraphics[width=0.7\columnwidth]{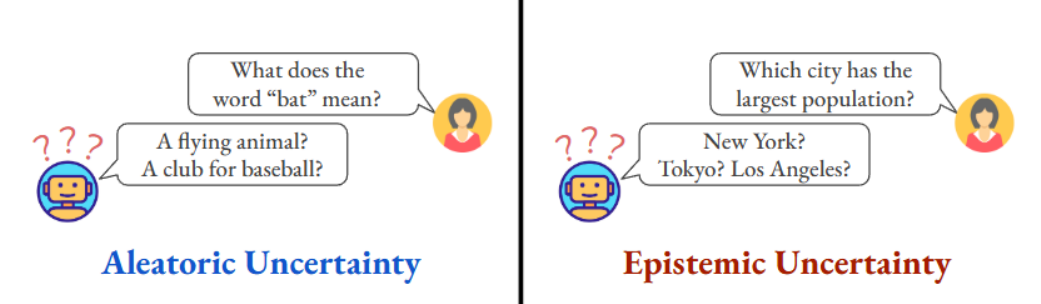}
\caption{Examples of aleatoric and epistemic uncertainty in natural language.}
\label{fig:uncertainty_ex}
\end{figure}

\subsection{Uncertainty Quantification for Classification}
UQ for classification in deep learning has been extensively studied, particularly from Bayesian perspective. In this setting, instead of relying on a single deterministic label, we require a predictive probability distribution over classes. Formally, for an input $x$, a classifier parametrized by $\theta$ outputs
\[
p_\theta(y \mid x), \quad y \in \{1, \ldots, K\},
\]
which assigns a probability to each class. This probabilistic formulation allows the model not only to identify the most likely class but also to assess the degree of confidence assigned to each alternative, thereby enabling principled quantification of uncertainty. 
A simple way to measure uncertainty from the predictive distribution is through the maximum predicted probability, which is defined as
\[
p_b(x) = \max_{c \in \{1,\dots,K\}} p_\theta(y=c \mid x),
\]
for a $K$-class classification problem.
It reflects the model’s most confident class assignment. The corresponding uncertainty score is then given by $U_{\mathrm{MP}}(x) = 1 - p_b(x),$ representing the degree of confidence the model assigns to its predicted class.

In this section, we highlight two representative perspectives that illustrate how uncertainty has been interpreted in classification: \textit{entropy of the posterior predictive distribution}, which provides a standard probabilistic framework, and \textit{pointwise Bayesian risk}, which frames uncertainty in terms of expected error. We then explain how these perspectives naturally lead to the uncertainty decomposition into aleatoric and epistemic components. Finally, we introduce approximation techniques that make UQ practically feasible in modern deep learning.

\noindent
\textbf{Uncertainty as Entropy of Posterior Predictive Distribution.}\quad
The entropy of the posterior predictive distribution is widely employed as a measure of predictive uncertainty~\cite{houlsby2011bayesianactivelearningclassification, schweighofer2024on}.
Given a dataset $\mathcal{D}$ and a prior $p(\mathbf{w})$ over model parameters, Bayes’ theorem provides the posterior distribution $p(\mathbf{w} \mid \mathcal{D})$. 
By integrating over the posterior, we obtain the posterior predictive distribution as
$p(y \mid \mathbf{x}, \mathcal{D}) = \mathbb{E}_{p(\mathbf{w} \mid \mathcal{D})}[p(y \mid \mathbf{x}, \mathbf{w})].$
Its entropy $H(p(y \mid x, \mathcal{D}))$ is taken as the overall measure of predictive uncertainty. 
Following~\cite{10.5555/1146355, 10.5555/3666122.3666976}, entropy can be decomposed as
\[
\underbrace{H(p(y \mid \mathbf{x}, \mathcal{D}))}_{\text{Total Uncertainty}}
= \underbrace{\mathbb{E}_{p(\mathbf{w} \mid \mathcal{D})}[H(p(y \mid \mathbf{x}, \mathbf{w}))]}_{\text{Aleatoric Uncertainty}}
+ \underbrace{I(y, \mathbf{w} \mid \mathbf{x}, \mathcal{D})}_{\text{Epistemic Uncertainty}}.
\]
The first expected entropy term is interpreted as aleatoric uncertainty, as it reflects the variability that persists even if the model parameters were known exactly, corresponding to randomness inherent in the data distribution.  
The second term, given by the mutual information of the posterior distribution, captures how much information about the model parameters $\mathbf{w}$ is revealed through the prediction of $y$, is therefore associated with epistemic uncertainty.

\noindent
\textbf{Uncertainty as Pointwise Bayes Risk.}\quad
Beyond the posterior-predictive perspective, Bayesian decision theory provides an alternative way to formalize predictive uncertainty by linking it to the concept of \textit{risk}~\cite{kotelevskii2025from, vashurin2025uncertaintyquantificationllmsminimum}.
Given a predictive distribution $\hat{p}_{\mathbf{w}}(y \mid \mathbf{x})$ and the true conditional distribution $p(y \mid \mathbf{x})$, the pointwise risk is defined as
\[
R(\hat{p}_{\mathbf{w}} \mid \mathbf{x}) 
= \sum_{k=1}^K L(\hat{p}_{\mathbf{w}}(y \mid \mathbf{x}), k) \, p(y=k \mid \mathbf{x}),
\]
where $L(\cdot)$ is a loss function, commonly zero-one loss or cross-entropy.  
This measures the expected error with input $\mathbf{x}$ under the true data-generating distribution and thus provides a natural way to connect predictive error with uncertainty.  
The pointwise risk can be decomposed as:
\[
R(\hat{p}_{\mathbf{w}} \mid \mathbf{x}) 
= \underbrace{R_{\mathrm{Bayes}}(\mathbf{x})}_{\text{Aleatoric Uncertainty}}
+ \underbrace{R_{\mathrm{Excess}}(\hat{p}_{\mathbf{w}} \mid \mathbf{x})}_{\text{Epistemic Uncertainty}}.
\]
Here, the Bayes risk is defined as:
\[
R_{\mathrm{Bayes}}(\mathbf{x}) 
= \sum_{k=1}^K L(p(y \mid \mathbf{x}), k)\, p(y=k \mid \mathbf{x}),
\]
which depends only on the true conditional distribution $p(y \mid \mathbf{x})$.  
It expresses the degree of ambiguity inherent in the data itself independent of the model’s parameters or architecture, therefore interpreted as aleatoric uncertainty.  
In contrast, the excess risk
\[
R_{\mathrm{Excess}}(\hat{p}_{\mathbf{w}} \mid \mathbf{x}) 
= R(\hat{p}_{\mathbf{w}} \mid \mathbf{x}) - R_{\mathrm{Bayes}}(\mathbf{x})
\]
captures the additional error arising from finite data, limited capacity, or model misspecification.  
It quantifies the part of predictive uncertainty attributable to lack of knowledge about the true data distribution and is thus interpreted as epistemic uncertainty.

\noindent \textbf{Practical Bayesian Approximations in Deep Learning.}\quad
The decomposition of predictive uncertainty into aleatoric and epistemic components provides a useful theoretical framework, but it cannot be directly applied in practice because the true data distribution is unknown. 
As a result, a range of approximation techniques have been proposed~\cite{he2025surveyuncertaintyquantificationmethods}. Bayesian neural networks approximate the posterior using techniques such as variational inference~\cite{Blei_2017, 10.1145/168304.168306, NIPS2011_7eb3c8be}, Laplace approximation~\cite{FRISTON2007220}, or Markov Chain Monte Carlo (MCMC)~\cite{10.1093/biomet/57.1.97}. 
More scalable alternatives include Monte Carlo dropout~\cite{pmlr-v48-gal16}, 
which applies dropout at inference time, resulting in different outputs across forward passes, and interprets this randomness as an approximation to Bayesian posterior sampling.
Another approach is deep ensembles~\cite{NIPS2017_9ef2ed4b, mallick2022deepensemblebaseduncertaintyquantificationspatiotemporal}, 
where multiple independently trained networks are combined, and the diversity across their predictions provides an empirical approximation of parameter uncertainty.
Finally, calibration methods such as temperature scaling adjust predictive probabilities to better align model confidence with empirical correctness. Together, these approaches demonstrate how Bayesian principles can be approximated in practice to provide useful estimates of uncertainty in deep learning.

\subsection{Uncertainty Quantification for LLMs}
UQ for LLMs presents unique challenges compared to that in deep learning classification tasks.
In classification, the model usually follows a closed-set assumption, where a fixed number of labels are known in advance, and the model’s confidence or uncertainty can be estimated from the output probability distribution over these labels.
In contrast, LLMs generate outputs in an open-ended space, where possible responses are virtually unbounded and cannot be reduced to a fixed label set.
Another source of complexity is that LLMs rely on an auto-regressive architecture, where the model generates one token at a time based on the conditional distribution of the next token given the history. 
Unlike classification where uncertainty is typically derived from a single probability distribution over a fixed label set, auto-regressive generation produces a new distribution at every decoding step. As a result, uncertainty cannot be directly defined using the same approaches as in classification, since it must be considered across an evolving sequence of token-level distributions rather than a single output distribution.

Another key distinction arises from the role of the input prompt. 
In classification tasks, the input typically maps to a well-defined label space, whereas in LLMs, the prompt not only provides the question or context but also determines the task to be performed. 
An LLM can be used for question answering (QA), translation, summarization, reasoning, or open-domain conversation depending entirely on how the prompt is phrased. 
Furthermore, even within the same task, the prompt can be expressed in multiple ways, leading to variations in the model’s response behavior. 
This flexibility greatly expands the range of possible outputs but also complicates UQ, as the model must manage uncertainty across diverse and implicitly defined tasks.

\noindent \textbf{Levels of Uncertainty.} \quad
Beyond explaining how UQ for LLM is different from UQ for classification tasks, we clarify what object uncertainty is defined over in the context of LLMs. 
One natural formulation is to define an uncertainty score $U(\mathbf{x}, \mathbf{y})$ for a query–answer pair, where $\mathbf{x}$ is the input prompt and $\mathbf{y}$ is the output generation. 
This pair-level definition is intuitive, as it evaluates how uncertain the model is about producing a specific response to a given query.
Alternatively, some methods assign an uncertainty score directly to a query $\mathbf{x}$, without conditioning on any specific generation. 
This query-level uncertainty score reflects how likely the model is to produce unreliable or inconsistent responses to input $\mathbf{x}$.

\subsection{Hallucination Problem}
Hallucination in LLMs refers to the generation of content that is factually incorrect or not grounded by the provided context. 
Such outputs hinder model reliability and raise safety concerns for real-world deployment, particularly in high-stakes domains such as healthcare, finance, law, or education. 
Prior work has categorized types of hallucination~\cite{Huang_2025}.

\noindent \textbf{Factuality Hallucination.}\quad
Factuality hallucination can be broadly categorized into \textit{factual contradiction} and \textit{factual fabrication}. 
Factual contradiction refers to cases where the model generates statements that conflict with real-world facts. 
Such contradictions can arise at any stage of the model’s use of factual knowledge, including storing, processing, and generating.
Factual contradiction includes cases where the model outputs incorrect entities, as well as cases where it generates incorrect relations between entities. 
In contrast, \textit{factual fabrication} refers to cases where the model invents information that does not exist in reality and presents it as if it were true. 
This includes cases where the model produces claims that are entirely unverifiable by any available sources, 
as well as exaggerated or overgeneralized outputs.

\noindent \textbf{Faithfulness Hallucination.}\quad
Faithfulness hallucination arises when a model’s outputs fail to remain consistent with the input prompt, the given context, or logical reasoning. 
From this perspective, it can be divided into three subtypes. 
\textit{Instruction inconsistency} occurs when the model’s response deviates from the prompt instruction, generating unintended outputs. 
\textit{Context inconsistency} refers to cases where the model relies on its own parametric knowledge and disregards or contradicts the provided contextual information. 
\textit{Logical inconsistency} arises when the model’s reasoning contains internal contradictions, either among intermediate steps or between the reasoning process and the final answer.

\subsection{Uncertainty Quantification for Hallucination Detection}

Recent studies show that LLMs tend to generate plausible but false answers when they are uncertain, rather than admitting their inability to provide valid outputs~\cite{kalai2025languagemodelshallucinate}.
Another line of evidence further suggests that low-confidence situations are more likely to produce hallucinations, as they generate answers regardless of a lack of knowledge or unstable reasoning.~\cite{huang2025confqaanswerconfident}. 
This motivates the use of UQ as natural basis for hallucination detection, since predictive uncertainty can serve as a useful signal for identifying unreliable outputs.
In practice, UQ is used as hallucination detector by setting a threshold on predictive uncertainty and making a binary decision about whether a given output should be flagged as a hallucinated output~\cite{bakman-etal-2025-reconsidering}. 

In terms of implementation, UQ can be realized through multiple approaches, including token-level probability and entropy measures, consistency checks across multiple outputs, probing of internal model states, and self-evaluation mechanisms. 
We provide a detailed classification of these methods in Section~\ref{sec:Category}.

\section{Classification of UQ Methods}
\label{sec:Category}

UQ methods can be distinguished along multiple dimensions, depending on how they approach uncertainty and how it is measured. 
In this section, we provide a systematic categorization of existing UQ methods and highlight the key insights of each category. 
Specifically, we introduce four independent axes of classification: 
\textit{conceptual approach}, \textit{sampling requirement}, \textit{model accessibility}, and \textit{training reliance}.
Every method has a position on each of the four axes, reflecting its specific characteristics across these dimensions. 
This framework captures the diversity of UQ methods more precisely and clarifies how different approaches relate to one another. 
Table~\ref{tab:ue-4axes} summarizes existing methods by positioning them along these axes, thereby providing a structured categorization of the current UQ literature.

\begin{figure*}[t]
\begin{center}
\includegraphics[width=1\textwidth]{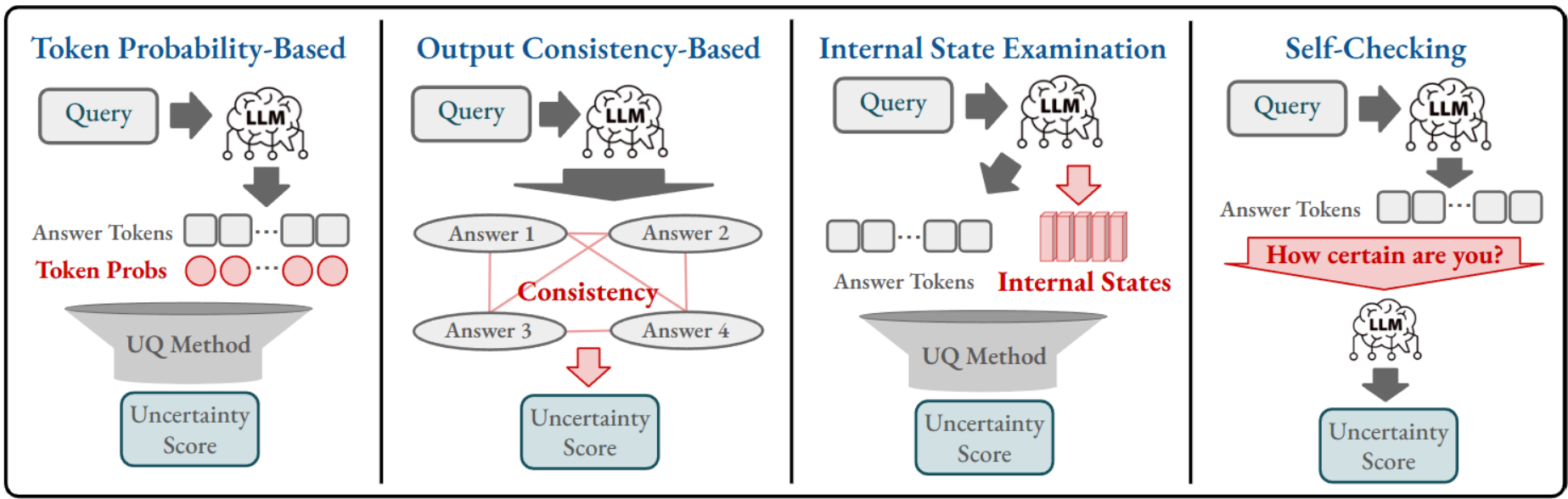}
\caption{Categorization of uncertainty quantification methods based on conceptual approaches.}
\label{fig:categorization}
\end{center}
\end{figure*}

\subsection{Conceptual Approaches}
\label{sec:concepts}
Uncertainty quantification can be approached through diverse conceptual perspectives, depending on what aspect of the model or its outputs is used to estimate uncertainty.
We classify UQ methods according to their conceptual approaches into four groups, as illustrated in Figure~\ref{fig:categorization}:  
(i) \textit{Token probability-based methods}, which estimate uncertainty directly from token probability distributions;  
(ii) \textit{Output consistency-based methods}, which assess uncertainty by examining the agreement or diversity among multiple outputs for a given input;  
(iii) \textit{Internal state examination methods}, which analyze intermediate model activations to detect potential errors; and  
(iv) \textit{Self-checking methods}, in which the model evaluates the correctness of its own generations using auxiliary prompts or strategies.  
While these groups provide a clear categorization, it is important to note that some methods may simultaneously belong to more than one category.  

\noindent \textbf{Token Probability-based Methods.}\quad
The most straightforward way to estimate uncertainty in LLMs is to utilize the probabilities of tokens in the generated sequence. 
A natural measure of confidence for a response $\mathbf{y}$ given a query $\mathbf{x}$ is the \textit{conditional sequence probability (CSP)}, defined as the log-probability of the entire generation:  
\begin{equation}
    \mathrm{CSP}(\mathbf{y}, \mathbf{x}) = \log P(\mathbf{y} \mid \mathbf{x}) 
    = \sum_{j=1}^{L} \log P(y_{j} \mid \mathbf{y}_{<j}, \mathbf{x}),
\end{equation}
where $L$ denotes the length of the generation and $\mathbf{y}_{<j}$ represents the previous tokens up to $j-1$. 
This formulation captures the intuition that a sequence with higher probability indicates lower uncertainty, serving as the foundation for a variety of methods that build directly on token-level probabilities.

\noindent \textbf{Output Consistency-based Methods.}\quad
The underlying intuition of output-consistency-based methods is that inconsistent responses to the same prompt indicate higher uncertainty.
Accordingly, multiple generations are sampled for a given query, and their degree of alignment is examined to estimate uncertainty.
In practice, such methods often measure the similarity between text fragments using NLI (natural language inference) scores to classify the relationship between two outputs as entailment, neutral, or contradiction. 
This principle provides a simple yet versatile foundation for methods that infer uncertainty from the variability of generated outputs.

\noindent \textbf{Internal State Examination Methods.}\quad
Uncertainty can also be estimated by analyzing internal signals from LLMs, such as hidden states, attention weights, or other intermediate representations that reflect the model’s internal processing.
\textit{Hidden states} refer to the embedding vectors from intermediate decoder layers for each generated token, reflecting how the model internally interprets the input during processing.
The intuition of using hidden states for UQ is that embeddings from hallucinated and factual responses occupy different regions in the feature space, enabling their separation. 
\textit{Attention weights} are the lower-triangular matrices that describe how each token attends to previous tokens during generation. 
Attention weight-based methods leverage the observation that the model attends differently when generating hallucinations compared to truthful responses, allowing attention patterns to be used as signals for uncertainty quantification. 
These internal representations provide rich signals for assessing uncertainty, uncovering how internal processing relates to generative reliability.

\noindent \textbf{Self-Checking Methods.}\quad
Self-checking approaches estimate uncertainty by directly querying the LLM about the reliability of its own generations. 
Instead of relying on auxiliary signals such as token probabilities or internal states, these methods depend solely on the model’s ability to assess the uncertainty of its own outputs.
In practice, this often involves prompting the model to justify or rate its answer, or to generate alternative answers for cross-validation. 
While highly flexible and applicable, this strategy may also inherit the model’s biases, making it less reliable. 
Nevertheless, self-checking provides a distinct paradigm that frames uncertainty quantification as a meta-reasoning task performed by the model itself.

\subsection{Sampling Requirement}
UQ methods can also be categorized by whether they require a single model output or multiple samples. 
Some methods require only one generation to assign an uncertainty score. As they require only one inference pass, these methods are computationally efficient and directly provide an uncertainty estimate for the specific output without reference to additional samples. 
Other approaches involve multiple outputs to quantify uncertainty.
They examine the collective behavior of sampled responses, such as diversity, variability or alignment among the outputs.
While this approach increases computational cost due to repeated inference, it often yields more reliable estimates, and using a larger number of samples can further improve the stability of the resulting scores.

\subsection{Model Accessibility (Black-box, Gray-box, White-box)}
Another important axis for categorizing uncertainty quantification methods is the level of access to the underlying model. 
We can broadly distinguish three settings: \textit{black-box}, \textit{gray-box}, and \textit{white-box}. 

\noindent \textbf{Black-box models.}\quad
In black-box scenarios, the model provides only its output generations without revealing any internal details, for security and proprietary reasons.
Restricted to only the responses without any further information, uncertainty can be quantified by analyzing output diversity across multiple generations or by prompting the model for self-assessment.

\noindent \textbf{Gray-box models.}\quad
Gray-box models provide limited access to internal signals beyond the raw output text. 
For example, they may expose token-level probabilities or logits, partial embedding representations, or other selectively available information. 
UQ methods in this setting leverage these additional signals to produce finer-grained estimates than black-box approaches, while still avoiding the heavy resource demands of full white-box access.  

\noindent \textbf{White-box models.}\quad
White-box scenarios enable a richer set of uncertainty quantification techniques that directly exploit the model’s intermediate representations, attention weights, or parameter-level information. 
While such approaches often provide deeper insights and stronger signals, they require hosting the model locally, which may not be available for all models.

\subsection{Training Reliance}
We categorize UQ methods in terms of their reliance on training: supervised methods and unsupervised methods. This distinction is crucial, as training-required methods introduce additional considerations such as the availability of training data including both inputs and labels, transferability across domains, and their adaptability across models.

\noindent \textbf{Supervised Methods.}\quad
Supervised UQ approaches rely on a labeled dataset and training a comparatively small model to quantify uncertainty.
Some of these approaches utilize synthetic datasets created by prompting LLMs to generate both faithful and hallucinated outputs. 
Features are then extracted from either the model’s internal states or its outputs, and used to train a supervised predictive model for estimating uncertainty.
The key advantage of supervised methods lies in their strong performance, as they can directly learn mappings from input features to uncertainty levels using labeled data.
In addition, they offer flexibility to design specialized uncertainty quantifiers that can be tailored to particular domains or tasks, provided that domain-specific training data is available.
However, their reliance on training data introduces challenges in terms of scalability and transferability across domains and models, as well as the practical difficulty of obtaining high-quality labeled datasets for uncertainty estimation.

\noindent \textbf{Unsupervised Methods.}\quad
In contrast, unsupervised approaches do not require additional training or labeled dataset. 
These training-free methods estimate uncertainty directly from the generative behavior of the LLM, for example by analyzing token probabilities or intrinsic signals, or examining the consistency of generated outputs.
As these methods avoid supervised training, unsupervised methods are less susceptible to dataset-specific biases and are generally more robust to distribution shifts. 
Moreover, they are often more computation-efficient, as they eliminate the need for dataset construction and model retraining. 
However, the main limitation is that these methods may provide less accurate or less task-specialized uncertainty estimates compared to supervised approaches.

\begin{table*}[t]
\centering
\resizebox{\textwidth}{!}{
\begin{tabular}{lcccc}
\toprule
\textbf{UQ Methods} & 
\shortstack{\textbf{Conceptual}\\\textbf{Approach}} & 
\shortstack{\textbf{Sampling}\\\textbf{Requirement}} & 
\shortstack{\textbf{Model Accessibility}\\(Black-, Gray-, White-box)} & 
\shortstack{\textbf{Training}\\\textbf{Reliance}} \\
\midrule
LNS~\cite{malinin2021uncertainty} & Token Prob. & Single & Gray-box & No \\
Entropy~\cite{malinin2021uncertainty} & Token Prob. & Multiple & Gray-box & No \\
Semantic Entropy~\cite{kuhn2023semantic} & Token Prob., Output Consistency & Multiple & Gray-box & No \\
SentSAR~\cite{tokensar} & Token Prob. & Multiple & Gray-box & No \\
MARS~\cite{bakman2024mars} & Token Prob. & Single & Gray-box & No \\
TokenSAR~\cite{tokensar} & Token Prob. & Single & Gray-box & No \\
SAR~\cite{tokensar} & Token Prob. & Multiple & Gray-box & No \\
LARS~\cite{yaldiz2024designlearntrainablescoring} & Token Prob. & Single & Gray-box & Yes \\
PMI~\cite{takayama-arase-2019-relevant} & Token Prob. & Single & Gray-box & No \\
CPMI~\cite{van-der-poel-etal-2022-mutual} & Token Prob. & Single & Gray-box & No \\
KLE~\cite{nikitin2024kernel} & Output Consistency & Multiple & Black-box & No \\
SumEigenV~\cite{lin2024generating} & Output Consistency & Multiple & Black-box & No \\
Degree Matrix~\cite{lin2024generating} & Output Consistency & Multiple & Black-box & No \\
Eccentricity~\cite{lin2024generating} & Output Consistency & Multiple & Black-box & No \\
Self-Detection~\cite{zhao-etal-2024-knowing} & Output Consistency & Multiple & Black-box & No \\
MI Estimator~\cite{abbasi-yadkori2024to} & Output Consistency & Multiple & Black-box & No \\
Semantic Density~\cite{qiu2024semantic} & Output Consistency & Multiple & Black-box & No \\
CoCoA~\cite{vashurin2025uncertaintyquantificationllmsminimum} & Token Prob., Output Consistency & Multiple & Gray-box & No \\
MD~\cite{DBLP:conf/nips/LeeLLS18} & Internal State Examination & Single & White-box & No \\
CSL~\cite{lin-etal-2024-contextualized} & Internal State Examination & Single & White-box & No \\
INSIDE~\cite{chen2024inside} & Internal State Examination & Multiple & White-box & No \\
Attention Scores~\cite{sriramanan2024llmcheck} & Internal State Examination & Single & White-box & No \\
SAPLMA~\cite{azaria-mitchell-2023-internal} & Internal State Examination & Single & White-box & Yes \\
Lookback Lens~\cite{chuang-etal-2024-lookback} & Internal State Examination & Single & White-box & Yes \\
Feature-Gaps~\cite{bakman2025uncertaintyfeaturegapsepistemic} & Internal State Examination & Single & White-box & Yes \\
Focus~\cite{zhang-etal-2023-enhancing-uncertainty} & Internal State Examination & Single & White-box & No \\
RAUQ~\cite{vazhentsev2025uncertaintyawareattentionheadsefficient} & Internal State Examination & Single & White-box & No \\
HaloScope~\cite{du2024haloscope} & Internal State Examination & Multiple & White-box & Yes \\
SEPs~\cite{kossen2024semanticentropyprobesrobust} & Internal State Examination & Multiple & White-box & Yes \\
UQAC~\cite{li2025language} & Internal State Examination & Single & White-box & No \\
Factoscope~\cite{he-etal-2024-llm} & Internal State Examination & Single & White-box & Yes \\
P(True)~\cite{kadavath2022language} & Self-Checking & Single & Black-box & No \\
Verbalized Confidence~\cite{tian-etal-2023-just} & Self-Checking & Single & Black-box & No \\
Cross-Examination~\cite{cohen-etal-2023-lm} & Self-Checking, Output Consistency & Multiple & Black-box & No \\
BS Detector~\cite{chen-mueller-2024-quantifying} & Self-Checking, Output Consistency & Multiple & Black-box & No \\
\bottomrule
\end{tabular}
}
\caption{UQ methods categorized along four axes: \textit{conceptual approach, sampling requirement, model accessibility, and training reliance.}}
\label{tab:ue-4axes}
\end{table*}

\section{Uncertainty Quantification Methods}
\label{sec:Methods}

In this section, we provide a brief overview of representative uncertainty quantification methods. 
For each approach, we highlight its core idea and main characteristics.
Here, we follow the categorization of the methods according to \textit{conceptual approaches} defined in Section~\ref{sec:concepts}, 
assigning each method to the category that best reflects its primary insight.

\subsection{Token Probability-based Methods}

\textbf{Length Normalized Scoring (LNS)} \cite{malinin2021uncertainty} computes the average log-probability of each token in the generated sequence:
\begin{equation}
     \log\tilde{P}(\mathbf{y}|\mathbf{x}, \theta) = \frac{1}{L} \sum_{l=1}^{L} \log P(y_l|\mathbf{y}_{<l}, \mathbf{x}; \theta),
\label{length-normalized-prob}
\end{equation}
where $P(\mathbf{y}|\mathbf{x}, \theta)$ represents the probability of the generated sequence $\mathbf{y}$ (of length $L$), and $\mathbf{y}_{<l} \triangleq \{y_1, y_2, \dots, y_{l-1}\}$ denotes the tokens generated before token $y_l$.
We note that \textbf{perplexity} measures the same quantity as LNS but in the exponential scale, and is defined as:
\begin{equation}
    \text{Perplexity}(\mathbf{y}) = \exp\!\left(-\frac{1}{L}\sum_{l=1}^{L} \log P(y_l \mid \mathbf{y}_{<l}, \mathbf{x}; \theta)\right).
\end{equation}

\textbf{Entropy} \cite{malinin2021uncertainty} estimates uncertainty by sampling multiple generations for a given query $\mathbf{x}$, computing the LNS for each sample, and averaging over them. This approach corresponds to a Monte Carlo approximation over the generation space:
\begin{equation}
\mathcal{H}(\mathbf{x},\theta) \approx - \frac{1}{B} \sum_{b =1}^{B} \log \tilde{P}(\mathbf{y}_b|\mathbf{x}, \theta),
\label{entropy}
\end{equation}
where $B$ represents the number of sampled generations.

\textbf{Semantic Entropy} \cite{kuhn2023semantic} refines entropy quantification by leveraging the semantic meanings of sampled generations. Instead of treating all generations equally, it clusters semantically equivalent responses and computes entropy based on the probability distribution over clusters:
\begin{equation}
\mathrm{SE}(\mathbf{x},\theta) = - \frac{1}{|C|} \sum_{i=1}^{|C|} \ln P(\mathrm{c}_i|\mathbf{x}, \theta),
\label{semantic-entropy}
\end{equation}
where $\mathrm{c}_i$ denotes a semantic cluster, and $C$ represents the set of all clusters. 

Similarly, \textbf{SentSAR} \cite{tokensar} computes pairwise similarities between generations and assigns higher entropy weights to sentences that are more similar to others. This method can be interpreted as a weighted version of Semantic Entropy. Instead of binary entailment decisions, SentSAR assigns a continuous similarity score to each sentence. 

\textbf{MARS} \cite{bakman2024mars} and \textbf{TokenSAR} \cite{tokensar} enhance entropy-based scoring by incorporating the contribution of individual tokens to the overall meaning. These approaches refine probability-based scoring by weighting token probabilities differently:
\begin{equation}
    \bar{P}(\mathbf{y}|\mathbf{x}, \theta) = \prod_{l=1}^{L} P(y_l|\mathbf{y}_{<l}, \mathbf{x}; \theta)^{w(\mathbf{y},\mathbf{x}, L, l)},
\label{mars}
\end{equation}
where $w(\mathbf{y}, \mathbf{x}, L, l)$ represents the token weight assigned by MARS or TokenSAR. These methods aim to emphasize tokens that directly contribute to answer the query (MARS) or are semantically significant (TokenSAR). \textbf{SAR} extends this approach by combining TokenSAR and SentSAR.

\textbf{LARS}~\cite{yaldiz2024designlearntrainablescoring} 
introduces a trainable scoring model. 
LARS employs an encoder-only transformer that takes as input the query, the generated response tokens, and their corresponding probabilities, and outputs a reliability score. 
To integrate probabilities into the same embedding space as text tokens, each token probability is discretized, allowing the model to jointly process linguistic and probabilistic signals within a unified sequence.

\textbf{Pointwise Mutual Information (PMI)} \cite{takayama-arase-2019-relevant} compares the likelihood of a token under the language model and under the source-conditioned model. 
For a generated sequence $\mathbf{y} = (y_1, \dots, y_L)$ given source $\mathbf{x}$, the mean PMI is defined as
\begin{equation}
\mathrm{PMI}(\mathbf{y},\mathbf{x};\theta) = \frac{1}{L} \sum_{l=1}^{L} 
\log \frac{P(y_l \mid \mathbf{y}_{<l},\theta)}{P(y_l \mid \mathbf{y}_{<l},\mathbf{x},\theta)},
\end{equation}
which reflects the degree to which the generated tokens are supported by the source.

\textbf{Conditional PMI (CPMI)} \cite{van-der-poel-etal-2022-mutual} 
extends PMI by incorporating model uncertainty measured through conditional entropy. 
When the entropy of the next-token distribution $H(y_l \mid y_{<l},\mathbf{x},\theta)$ exceeds a threshold $\tau$, 
the uncertainty is measured as
\begin{align}
\mathrm{CPMI}(\mathbf{y},\mathbf{x};\theta) = 
& - \frac{1}{L} \sum_{l=1}^{L} 
\log P(y_l \mid \mathbf{y}_{<l},\mathbf{x},\theta) \notag \\
& + \frac{\lambda}{L} \sum_{l: H(y_l \mid \mathbf{y}_{<l},\mathbf{x},\theta) \geq \tau} 
\log P(y_l \mid \mathbf{y}_{<l},\theta),
\end{align}
where $\lambda > 0$ is a tunable parameter controlling the influence of the marginal probability.

\subsection{Output Consistency-based Methods}

\textbf{Kernel Language Entropy (KLE)} \cite{nikitin2024kernel} quantifies uncertainty using the von Neumann entropy (VNE) of the semantic kernel $K_{sem}$, which is constructed from LLM generations $\mathbf{y_1}, \dots, \mathbf{y_N}$ and the input $\mathbf{x}$:
\begin{equation}
KLE(x) = VNE(K_{sem}).
\end{equation}
A semantic graph is first defined where edges encode pairwise entailment dependencies between output sequences:
\begin{equation}
W_{ij} = f(NLI(\mathbf{y_i},\mathbf{y_j}), NLI(\mathbf{y_j},\mathbf{y_i})).
\end{equation}
The graph Laplacian is computed as $L = D - W$, where the degree matrix $D$ is defined as:
\begin{equation}
D_{ii} = \sum_{j=1}^{|V|}W_{ij}.
\end{equation}
Then, a heat kernel $K_t = e^{-tL}$. To obtain a unit-trace positive semidefinite kernel, we apply the following normalization:
\begin{equation}
K(x,y) \leftarrow K(x,y) (K(x,x)K(y,y))^{-1/2} / N,
\end{equation}
where $N$ is the size of $K$. Finally, the kernel entropy is computed using the von Neumann entropy (VNE):
\begin{equation}
VNE(A) \triangleq -\text{Tr}[A\log A].
\end{equation}

\textbf{SumEigenV} \cite{lin2024generating} is computed using the Laplacian matrix $L$:
\begin{equation}
L \triangleq I - D^{-\frac{1}{2}} W D^{-\frac{1}{2}}.
\end{equation}
The final SumEigenV score is defined as:
\begin{equation}
\text{SumEigV} = \sum_{k=1}^{N} \max(0, 1 - \lambda_k),
\end{equation}
where $\lambda_1, \dots, \lambda_N$ are the eigenvalues of the Laplacian matrix $L$.
Using the same degree matrix $D$, we define \textbf{Degree Matrix Uncertainty} and \textbf{Degree-Matrix-C} \cite{lin2024generating} for a given generation $j$ as:
\begin{flalign}
&\text{Degree Matrix Uncertainty} = \frac{\text{trace}(mI - D)}{m^2}, \\
&\text{Degree Matrix-C} = \frac{D_{j,j}}{m}.
\end{flalign}

\textbf{Eccentricity Uncertainty} and \textbf{Eccentricity-C} \cite{lin2024generating} are computed as follows. First, we obtain the smallest $k$ eigenvectors, $\mathbf{u_1}, \dots, \mathbf{u_k}$. For each generation $j$, we construct the vector $\mathbf{v_j} = [u_{1,j}, ...,  u_{k,j}]$. Then, the uncertainty measures are defined as:
\begin{equation}
\begin{aligned}
\text{Eccentricity Uncertainty} &= \left\| \begin{bmatrix} \mathbf{v}_1'^{\top}, \dots, \mathbf{v}_N'^{\top} \end{bmatrix} \right\|_2, \\
\text{Eccentricity-C}  &= -\|\mathbf{v}_j'\|_2.
\end{aligned}
\end{equation}
where $\mathbf{v}_j' = \mathbf{v}_j - \frac{1}{m} \sum_{j'=1}^{m} \mathbf{v}_{j'}$.

\textbf{Self Detection} \cite{zhao-etal-2024-knowing} paraphrases each question five times and clusters the generations based on entailment relationships. An entropy score is then computed over these clusters as follows:
\begin{equation}
\text{Self Detection Entropy} = - \sum_{c_i \in C} \frac{|c_i|}{N_q} \ln\left(\frac{|c_i|}{N_q}\right),
\end{equation}
where $C$ represents the set of clusters and $N_q$ is the number of paraphrased questions.

\textbf{Mutual Information (MI) Estimator} \cite{abbasi-yadkori2024to} quantifies epistemic uncertainty by computing the mutual information between the empirical joint distribution of model outputs $\tilde{Q}$ and the product of its marginals $\tilde{Q}^{\otimes}$:
\begin{equation}
I(\tilde{Q}) = D_{\mathrm{KL}}(\tilde{Q} \,\|\, \tilde{Q}^{\otimes}).
\end{equation}
The estimator replaces the true distribution with an empirical distribution $\hat{\mu}$, constructed from these sampled generations, along with its product of marginals $\hat{\mu}^{\otimes}$. 
The mutual information is then approximated as
\begin{equation}
\hat{I}_k(\gamma_1, \gamma_2) = \sum_{i \in U} \hat{\mu}(X_i) \ln \frac{\hat{\mu}(X_i) + \gamma_1}{\hat{\mu}^{\otimes}(X_i) + \gamma_2},
\end{equation}
where $U$ denotes the set of unique sampled outputs, and $(\gamma_1,\gamma_2)$ are smoothing terms that account for unseen events and ensure stability.

\textbf{Semantic Density (SD)} \cite{qiu2024semantic} 
quantifies confidence by estimating how densely a response is supported in semantic space. 
Given a prompt $\mathbf{x}$ and a target response $\mathbf{y}^{*}$ with embedding $\mathbf{v}^{*} = E(\mathbf{y}^{*}|\mathbf{x})$, 
a set of $M$ reference responses $\{\mathbf{y}_i\}_{i=1}^{M}$ with embeddings $\mathbf{v}_i = E(\mathbf{y}_i|\mathbf{x})$ 
is sampled from the LLM. 
The semantic density of $\mathbf{y}^{*}$ is defined as
\begin{equation}
\mathrm{SD}(\mathbf{y}^{*}|\mathbf{x}) = 
\frac{1}{\sum_{i=1}^{M} p(\mathbf{y}_i|\mathbf{x})} 
\sum_{i=1}^{M} p(\mathbf{y}_i|\mathbf{x}) \, K(\mathbf{v}^{*} - \mathbf{v}_i),
\end{equation}
where $p(\mathbf{y}_i|\mathbf{x})$ denotes the sequence probability of $\mathbf{y}_i$ 
and $K(\cdot)$ is a kernel function measuring semantic similarity. 
A higher density indicates that $\mathbf{y}^{*}$ lies close to other high-probability responses, thus implying higher confidence.

\textbf{CoCoA}~\cite{vashurin2025uncertaintyquantificationllmsminimum} 
quantifies uncertainty by combining token-probability-based confidence with output-consistency-based semantic similarity. 
Confidence is measured as $1 - p(\mathbf{y}|\mathbf{x})$, where $p(\mathbf{y}|\mathbf{x})$ is the model-assigned probability of the output sequence, 
while consistency is estimated from the semantic similarity between the evaluated sequence and multiple sampled generations. 
Given $M$ sampled generations $\{\mathbf{y}^{(i)}\}_{i=1}^M$, the uncertainty of an evaluated output $\mathbf{y}^{*}$ is defined as
\begin{equation}
\hat{U}_{\text{CoCoA}}(\mathbf{y}^{*} | \mathbf{x}) 
= \big(1 - p(\mathbf{y}^{*}|\mathbf{x})\big) \cdot \frac{1}{M} \sum_{i=1}^{M} \big(1 - s(\mathbf{y}^{*}, \mathbf{y}^{(i)})\big),
\label{cocoa}
\end{equation}
where $s(\mathbf{y}^{*}, \mathbf{y}^{(i)}) \in [0,1]$ represents a semantic similarity function between outputs.

\subsection{Internal State Examination Methods}

\textbf{Mahalanobis Distance (MD)} \cite{DBLP:conf/nips/LeeLLS18} 
estimates uncertainty by measuring how far a test representation lies from the distribution of training representations. 
Let $\mathbf{h}(\mathbf{x})$ denote the hidden representation of an input $\mathbf{x}$. 
Assuming a Gaussian distribution over the training representations with empirical mean vector $\boldsymbol{\mu}$ and empirical covariance matrix $\boldsymbol{\Sigma}$, the Mahalanobis distance is defined as
\begin{equation}
    \mathrm{MD}(\mathbf{x}) = 
    \big(\mathbf{h}(\mathbf{x}) - \boldsymbol{\mu} \big)^{\top} 
    \boldsymbol{\Sigma}^{-1} 
    \big(\mathbf{h}(\mathbf{x}) - \boldsymbol{\mu} \big).
\end{equation}

\textbf{Contextualized Sequence Likelihood (CSL)} \cite{lin-etal-2024-contextualized} refines the standard sequence likelihood by reweighting token probabilities with attention-derived importance scores. 
Given a generated sequence $\mathbf{y} = (y_1, \dots, y_n)$ and input $\mathbf{x}$, 
the contextualized score is defined as
\begin{equation}
\mathrm{CSL}(\mathbf{y}, \mathbf{x}) = \sum_{i=1}^{n} w_i \, l_i, 
\quad w_i = \frac{a_i}{\sum_{j=1}^{n} a_j},
\label{csl}
\end{equation}
where $l_i = \log \hat{p}(y_i \mid \mathbf{y}_{<i}, \mathbf{x})$ is the logit of token $y_i$, 
and $a_i$ denotes the attention weight assigned to $y_i$.

\textbf{INSIDE} \cite{chen2024inside} originally composed of two main parts: EigenScore and test time feature clipping. The former one manipulates the activation of each new token during the generation process, which we do not include in our implementation. EigenScore calculates the semantic divergence in the hidden states of the model over sampled generations. First, for $B$ sampled generations, a covariance matrix is created $\Sigma = \mathbf{Z}^T \cdot \mathbf{J} \cdot \mathbf{Z}$. Here, each column of $\mathbf{Z}$ is the middle layer hidden state of the last token a sampled generation, and $J = I_d - \frac{1}{d} 1_d 1_d^T$, while $d$ being the hidden dimension. Then, the uncertainty score is calculated as follows:
\begin{equation}
\text{Inside}(x, \theta) = \frac{1}{B} \sum_{i} \log(\lambda_i)
\end{equation}
where $\lambda_i$'s are the eigenvalues of the regularized covarience matrix $\Sigma + \alpha I_K$.

\textbf{Attention Scores} \cite{sriramanan2024llmcheck} compute the log-determinant of the attention matrices across all heads of selected layers and sum them. This computation can be efficiently performed by summing the logarithm of the diagonal elements of each attention kernel:
\begin{equation}
-\log \det (Ker_i) = -\sum_{j=1}^{m} \log Ker_i^{jj},
\end{equation}
where $Ker_i$ represents the attention kernel matrix of head $i$.

\textbf{SAPLMA} \cite{azaria-mitchell-2023-internal} builds on the observation that hidden layers encode signals about whether a language model internally treats a statement as true or false, and trains an MLP-based classifier on hidden representations $\mathbf{h}(\mathbf{x})$ from a selected LLM layer to predict uncertainty of an input query $\mathbf{x}$.
\begin{equation}
F(\mathbf{h}(\mathbf{x})) = \sigma(\mathbf{W}^{\top}\mathbf{h}(\mathbf{x}) + b),
\end{equation}
The classifier is trained on a dataset of true/false sentences spanning multiple topics.

\textbf{Lookback Lens} \cite{chuang-etal-2024-lookback} detects uncertainty by leveraging attention distributions within the model. 
At each decoding step $t$, for head $h$ in layer $l$, the lookback ratio is defined as
\begin{equation}
LR^{l,h}_t = 
\frac{A^{l,h}_t(\text{context})}
{A^{l,h}_t(\text{context}) + A^{l,h}_t(\text{new})},
\end{equation}
where $A^{l,h}_t(\text{context})$ and $A^{l,h}_t(\text{new})$ denote the average attention weights assigned to the input context tokens and the newly generated tokens, respectively. 
The lookback ratios across all heads and layers are concatenated into a vector $\mathbf{v}_t$, and averaged over a span to form the aggregated lookback ratio vector $\bar{\mathbf{v}}$. 
A logistic regression classifier $F$ is trained on $\bar{\mathbf{v}}$ to predict whether the span is factual or hallucinated:
\begin{equation}
F(\bar{\mathbf{v}}) = \sigma(\mathbf{w}^{\top}\bar{\mathbf{v}} + b).
\end{equation}

\textbf{Feature-Gaps}~\cite{bakman2025uncertaintyfeaturegapsepistemic} 
computes epistemic uncertainty by approximating the hidden-state distance between the given model and an ideal  model through a small set of semantically interpretable features. 
Specifically, the epistemic uncertainty term is upper-bounded by the norm of their hidden-state difference, 
$U_{\text{epi}} \leq 2\|W\|\,\|h_t^* - h_t\|$, 
which is further expressed as a composition of \textit{feature gaps} under the linear representation hypothesis:
\begin{equation}
\|h_t^* - h_t\| = 
\Big\|
\sum_{v_i \in \mathcal{H}} (\beta_i - \alpha_i) v_i
\Big\|,
\end{equation}
where $\mathcal{H}$ denotes a subset of latent semantic features that best approximate the model–ideal discrepancy. 
For contextual QA, three high-level features are used: 
\textit{context reliance}, \textit{context comprehension}, and \textit{honesty}.
Each feature vector $v_i$ is extracted from contrastive prompts (e.g., ``use your own knowledge'' vs. ``look at the context'') and identified through PCA. 
Finally, the epistemic uncertainty score is computed as a linear combination of the three feature representations, where each feature’s contribution is scaled by a learned coefficient.

\textbf{Focus} \cite{zhang-etal-2023-enhancing-uncertainty} 
detects hallucinations by refining token-level uncertainty scores and aggregating them with a stronger focus on informative tokens. 
For each generated token $t_i$, the uncertainty score is computed as
\begin{equation}
h_i = - \log p_i(t_i) + H_i, 
\quad H_i = - \sum_{v \in V} p_i(v) \log p_i(v),
\end{equation}
where $p_i(v)$ denotes the probability of generating token $v$ at position $i$ over the vocabulary $V$. 
As hallucination propagates to subsequent tokens, token-level uncertainty is adjusted as
\begin{equation}
h'_j = h_j + \sum_{i < j} A_{ji} \, h_i,
\end{equation}
where $A_{ji}$ is the attention weight from token $t_j$ to $t_i$.
Next, token scores are reweighted according to their estimated importance $
\tilde{h}_i = w_i \, h'_i,$
where $w_i$ reflects token properties such as entity type or inverse document frequency, assigning higher weights to semantically salient or rare tokens. 
Finally, the sentence-level hallucination score is obtained by aggregating the adjusted token scores 
$
H(r) = \frac{1}{N} \sum_{i=1}^{N} \tilde{h}_i,
$
where $N$ is the number of tokens in the response $r$.

\textbf{Recurrent Attention-based Uncertainty Quantification (RAUQ)}~\cite{vazhentsev2025uncertaintyawareattentionheadsefficient} is an unsupervised method that estimates uncertainty by weighting token probabilities with attention signals from selected \textit{uncertainty-aware} heads and accumulating them recurrently across the sequence.
For each generated sequence $\mathbf{y} = (y_1, \dots, y_N)$ given input $\mathbf{x}$, 
a recurrent confidence score is computed as
\begin{equation}
c_l(y_i) =
\begin{cases}
P(y_i \mid \mathbf{x}), & i = 1, \\
\alpha \, P(y_i \mid y_{<i}, \mathbf{x}) + (1-\alpha) \, a^{l,h_l}_{i,i-1} \, c_l(y_{i-1}), & i > 1,
\end{cases}
\end{equation}
where $a^{l,h_l}_{i,i-1}$ is the attention weight from token $y_i$ to $y_{i-1}$ in the selected head $h_l$ of layer $l$, and $\alpha$ is a tunable parameter.  
The layer-wise sequence uncertainty is then defined as
\begin{equation}
u_l(\mathbf{y}) = - \frac{1}{N} \sum_{i=1}^{N} \log c_l(y_i),
\end{equation}
and the final uncertainty score aggregates across layers by
\begin{equation}
u(\mathbf{y}) = \max_{l \in L} u_l(\mathbf{y}),
\end{equation}
where $L$ denotes the set of informative layers.

\textbf{HaloScope} \cite{du2024haloscope} estimates uncertainty for hallucination detection by separating reliable and hallucinated generations in the representation space of LLMs. 
Given an unlabeled mixture distribution of truthful and hallucinated responses, embeddings $\mathbf{f}_i$ are extracted from the LLM and factorized using singular value decomposition (SVD). 
Then membership estimation score is then computed as the projection norm of $\mathbf{f}_i$ 
onto the top-$k$ singular vectors:
\begin{equation}
\zeta_i = \frac{1}{k} \sum_{j=1}^{k} \sigma_j \cdot \langle \mathbf{f}_i, \mathbf{v}_j \rangle^2,
\end{equation}
where $\mathbf{v}_j$ and $\sigma_j$ are the $j$-th singular vector and singular value, respectively. 
Since hallucinated generations tend to exhibit larger variance in the embedding space, the leading singular vectors are assumed to capture hallucination-related directions, which indicates that a higher score indicates stronger alignment with the hallucination subspace. 
Using these scores, samples are partitioned into pseudo-labeled sets 
(truthful vs. hallucinated), and a binary classifier is trained with 
embeddings $\mathbf{f}_i$ as input and the pseudo-labels derived from $\zeta_i$ as labels.

\textbf{Semantic Entropy Probes (SEPs)} \cite{kossen2024semanticentropyprobesrobust} 
approximate semantic uncertainty by training lightweight probes on hidden states of LLMs. 
Given an input query $\mathbf{x}$, the model generates a response and the hidden state 
$\mathbf{h}^l_p(\mathbf{x})$ from layer $l$ and token position $p$ is extracted. 
These representations are paired with semantic entropy scores $H_{\mathrm{SE}}(\mathbf{x})$ 
computed from sampled generations. 
A linear classifier $F$ is trained to predict the level of semantic entropy:
\begin{equation}
F(\mathbf{h}^l_p(\mathbf{x})) = \sigma(\mathbf{w}^\top \mathbf{h}^l_p(\mathbf{x}) + b),
\end{equation}
where the supervision signal is derived from binarized semantic entropy values.

\textbf{Uncertainty Quantification with Attention Chain (UQAC)}~\cite{li2025language} 
is designed to address the overconfidence problem that arises in long reasoning chains of LLMs. 
It quantifies uncertainty by tracing semantically crucial tokens in the reasoning process. 
Given an instruction sequence $\mathbf{x}_{\text{instr}}$, a response is decomposed into a reasoning sequence $\mathbf{x}_{\text{cot}}$ and a final answer $\mathbf{x}_{\text{ans}}$. 
UQAC constructs an attention chain $\mathbf{x}_{\text{attn}} \subset \mathbf{x}_{\text{cot}}$ by backtracking attention weights from $\mathbf{x}_{\text{ans}}$. 
The chain is refined through similarity filtering and probability thresholding to obtain a reduced set $S$ of reasoning paths. 
The model confidence is then approximated by marginalizing over $S$:
\begin{equation}
\tilde{P}(\mathbf{x}_{\text{ans}}|\mathbf{x}_{\text{instr}}) 
= \sum_{\mathbf{x}'_{\text{attn}} \sim S} P(\mathbf{x}_{\text{ans}}, \mathbf{x}'_{\text{attn}} | \mathbf{x}_{\text{instr}}).
\end{equation}

\textbf{Factoscope}~\cite{he-etal-2024-llm} 
detects factuality by extracting intermediate signals from LLMs and learning to distinguish factual from non-factual patterns. 
Given an input sequence $\mathbf{x}$ and its generated output $\mathbf{y}$, 
the method collects activation maps $A$, final output ranks $R$, 
top-$k$ output indices $T_k$, and top-$k$ output probabilities $P_k$ 
across layers.
These signals are transformed into embeddings 
through sub-models and integrated into a mixed representation $\mathbf{E}_{\text{mixed}}$. 
A Siamese network is then trained with triplet margin loss to separate factual from non-factual outputs. 
The factuality score for a test output is determined by nearest-neighbor comparison with a support set:
\begin{equation}
\hat{y} = \arg\min_{i} \; \mathrm{Dist}(\mathbf{E}_{\text{test}}, \mathbf{E}_{\text{sup}_i}),
\end{equation}
where $\mathbf{E}_{\text{test}}$ is the mixed embedding of the test output 
and $\mathbf{E}_{\text{sup}_i}$ are embeddings from the labeled support set.

\subsection{Self-Checking Methods}
\textbf{Ptrue}~\cite{kadavath2022language} measures uncertainty by evaluating the probability assigned to the tokens corresponding to the words “True” and “False” when the model is asked whether its response is correct.

\textbf{Verbalized Confidence}~\cite{tian-etal-2023-just} prompts the model to explicitly state its confidence in the correctness of a response as a numerical score between 0 and 100 for a given question-response pair.

\textbf{Cross-Examination}~\cite{cohen-etal-2023-lm} proposes a framework where separate LLMs take on different roles as an \textit{Examinee} and an \textit{Examiner}. 
Given a query, the examinee produces a claim, and the examiner is prompted to ask questions in multi-turn interactions in order to reveal contradiction between the responses.
Through this process, factual errors in the original claim can be detected based on inconsistencies identified during the examination.

\textbf{BS Detector}~\cite{chen-mueller-2024-quantifying} 
quantifies uncertainty by combining observed consistency across multiple sampled responses with self-reflection certainty elicited from the model itself. 
Given an input $\mathbf{x}$ and a generated answer $\mathbf{y}$, 
$k$ additional responses $\{\mathbf{y}_i\}_{i=1}^k$ are sampled from the LLM. 
Observed consistency $O$ is computed by measuring semantic similarity between 
$\mathbf{y}$ and each $\mathbf{y}_i$, using both natural language inference (NLI) contradiction scores and an indicator function: 
\begin{equation}
o_i = \alpha s_i + (1-\alpha) r_i, \quad 
O = \frac{1}{k} \sum_{i=1}^{k} o_i,
\end{equation}
where $s_i$ is the NLI-based similarity, 
$r_i = \mathds{1}[\mathbf{y}=\mathbf{y}_i]$, and $0 \leq \alpha \leq 1$.
Self-reflection certainty $S$ is obtained by prompting the LLM with follow-up questions about the correctness of $\mathbf{y}$ and mapping its categorical answers \{Correct, Incorrect, Not sure\} to numerical scores \{1.0, 0.0, 0.5\}. 
The final confidence score is a weighted combination of the two:
\begin{equation}
C(\mathbf{x}, \mathbf{y}) = \beta O + (1-\beta) S,
\end{equation}
where $0 \leq \beta \leq 1$. 

\section{Evaluating UQ Performance}
\label{sec:Eval}

Broadly, evaluation of UQ methods is conducted with respect to the correctness of the generated output. 
Given that the purpose of UQ is to differentiate reliable from unreliable outputs, most evaluation protocols compare model predictions against groundtruth answers and judge the quality of uncertainty scores based on how well they align with correctness labels. 
As correctness reflects the model's ability to generate the valid output, evaluations based on correctness inherently target epistemic uncertainty.
Accordingly, this evaluation scheme is basically measuring the performance of epistemic uncertainty quantification. 

Formally, given an input query $\mathbf{x}$, a model output $\mathbf{y}$, and a reference answer $\mathbf{y}^*$, 
a UQ method assigns a score $UQ(\mathbf{x}, \mathbf{y})$ that should ideally 
be negatively correlated with the correctness of $\mathbf{y}$ relative to $\mathbf{y}^*$. 
To evaluate whether these scores are meaningful, each prediction is paired with a correctness label derived from comparing $\mathbf{y}$ to $\mathbf{y}^*$, and the resulting score–label pairs are aggregated. 
More formally, our objective is to maximize
\[
\mathbb{E}\Big[ \mathds{1}_{U(\mathbf{x}_1,\mathbf{y}_1) < U(\mathbf{x}_2,\mathbf{y}_2)} \cdot \mathds{1}_{\mathbf{y}_1 \in Y_1 \wedge \mathbf{y}_2 \notin Y_2} \Big],
\]
where $(\mathbf{x}_1, \mathbf{y}_1), (\mathbf{x}_2, \mathbf{y}_2) \sim D_{\text{test}}$, with $D_{\text{test}}$ denoting the evaluation dataset.
This expectation enforces ranking consistency: correct outputs should receive lower uncertainty scores than incorrect outputs, making high-uncertainty generations more likely to be wrong.

\subsection{Correctness Measures}
Evaluating correctness in LLMs is far from straightforward. 
Unlike conventional classification settings where predictions are discrete, LLM outputs are free-form and can be semantically correct even when they are lexically different from the groundtruth answers, making correctness assessment more complex. 
Without an appropriate correctness measure, uncertainty scores cannot be meaningfully aligned with model correctness.

A variety of approaches have been proposed to determine correctness of the generated output. Classical methods rely on string-based metrics such as Exact Match (EM), ROUGE~\cite{lin-2004-rouge}, and BLEU~\cite{papineni-etal-2002-bleu}, which compare lexical overlap between the generated and reference answers. 
While these measures are computationally efficient, they often fail to capture semantic equivalence. 
More recent approaches use LLMs themselves as evaluators, often referred to as \textit{LLM-as-a-Judge}. 
In this setting, an LLM is prompted with the input query, the generated answer, the reference answer, and optionally the supporting context, and asked to output a correctness judgment as a binary label (correct vs. incorrect) or a graded score. 
This approach provides greater flexibility in handling paraphrasing and semantic variation, although it may be biased by the judge model and requires extra computation.

\subsection{Performance Metrics}
Using the correctness measures discussed above, the performance of a UQ method can be evaluated using two broad categories of metrics: \textit{threshold-independent} and \textit{threshold-dependent} measures. Threshold-independent metrics are generally preferred in the literature, as they do not require selecting a method-specific cutoff and therefore allow for more robust comparisons across models and datasets. 
In contrast, threshold-dependent metrics rely on fixed thresholds and these decisions can vary depending on how the uncertainty scores are scaled.

\noindent \textbf{Threshold-independent metrics.} \quad
The most widely used threshold-free metrics are the Area Under the Receiver Operating Characteristic Curve (AUROC), the Prediction–Rejection Ratio (PRR), and the Area Under the Precision-Recall Curve (AUPRC)~\cite{malinin-etal-2017-incorporating, bakman-etal-2025-reconsidering, vashurin2025benchmarkinguncertaintyquantificationmethods}. 
AUROC evaluates a method’s ability to discriminate between correct and incorrect outputs across all possible thresholds, with values ranging from 0.5 (random performance) to 1.0 (perfect discrimination). 
A higher AUROC indicates that the UQ method consistently assigns higher confidence to correct outputs than to incorrect ones. 

PRR quantifies the relative precision gain obtained by rejecting low-confidence predictions, measuring how much precision improves as increasingly uncertain outputs are discarded.
Formally, it is defined as the ratio between the area under the rejection curve of the evaluated uncertainty scores and that of a random baseline, normalized by the gap between an oracle strategy and the same random baseline:  
\[
\text{PRR} = \frac{\text{AUC}_{\text{unc}} - \text{AUC}_{\text{rand}}}{\text{AUC}_{\text{oracle}} - \text{AUC}_{\text{rand}}},
\]
where $\text{AUC}_{\text{unc}}$ corresponds to the area under the precision-rejection curve for the given method, $\text{AUC}_{\text{oracle}}$ denotes the ideal uncertainty that perfectly orders outputs by correctness, and $\text{AUC}_{\text{rand}}$ represents random rejection. 
PRR values range from 0.0 (equivalent to random rejection) to 1.0 (matching the oracle).
A higher PRR indicates that the UQ method is more effective at filtering out unreliable outputs while retaining high-precision predictions.

AUPRC measures how well the methods prioritize correct outputs over incorrect ones.
It captures the trade-off between retaining more outputs and keeping those outputs reliable. A higher AUPRC means that as more responses are included, the proportion of correct ones remains consistently high. 
Thus, in the context of UQ, a high AUPRC indicates that the method is effective at filtering out unreliable generations while still preserving a broad set of correct answers.

\noindent \textbf{Threshold-dependent metrics.} \quad
Conventional classification metrics such as accuracy, precision, recall, and F1 score can also be applied to UQ evaluation. 
These metrics are computed by using a fixed threshold on uncertainty scores, which determines the binary correctness predictions.
While informative, their dependence on threshold selection introduces potential bias and makes direct comparison between methods challenging. 
For this reason, calibration is often required to align thresholds across different methods.

\begin{table*}[t]
\centering
\resizebox{\textwidth}{!}{%
\begin{tabular}{l|cc|cc|cc||cc|cc|cc}
\toprule
 & \multicolumn{6}{c||}{\textbf{LLaMA-3 8B}} & \multicolumn{6}{c}{\textbf{GPT-4o-mini}} \\
\cmidrule{2-13}
 & \multicolumn{2}{c|}{TriviaQA} & \multicolumn{2}{c|}{GSM8K} & \multicolumn{2}{c||}{FactScore-Bio}
 & \multicolumn{2}{c|}{TriviaQA} & \multicolumn{2}{c|}{GSM8K} & \multicolumn{2}{c}{FactScore-Bio} \\
 \textbf{UQ Methods}& \textbf{AUROC} & \textbf{PRR} & \textbf{AUROC} & \textbf{PRR} & \textbf{AUROC} & \textbf{PRR}
 & \textbf{AUROC} & \textbf{PRR} & \textbf{AUROC} & \textbf{PRR} & \textbf{AUROC} & \textbf{PRR} \\
\midrule
LARS~\cite{yaldiz2024designlearntrainablescoring}         & 0.861 & 0.783 & 0.834 & 0.719 & 0.677 & 0.391 & 0.852 & 0.766 & 0.840 & 0.686 & 0.640 & 0.294 \\
MARS~\cite{bakman2024mars}     & 0.763 & 0.635 & 0.730 & 0.488 & 0.660 & 0.367 & 0.792 & 0.668 & 0.735 & 0.480 & 0.655 & 0.405 \\
Self-Detection~\cite{zhao-etal-2024-knowing} & 0.780 & 0.590 & 0.556 & 0.090 & 0.687 & 0.369 & 0.799 & 0.587 & 0.736 & 0.421 & 0.671 & 0.313 \\
P(True)~\cite{kadavath2022language}    & 0.727 & 0.485 & 0.654 & 0.307 & 0.670 & 0.368 & 0.772 & 0.509 & 0.833 & 0.636 & 0.658 & 0.372 \\
Attention Scores~\cite{sriramanan2024llmcheck}   & 0.523 & 0.092 & 0.503 & -0.024 & 0.644 & 0.263 & -- & -- & -- & -- & -- & -- \\
Cross-Examination~\cite{cohen-etal-2023-lm} & 0.664 & 0.377 & 0.585 & 0.187 & 0.683 & 0.361 & 0.718 & 0.483 & 0.768 & 0.551 & 0.635 & 0.289 \\
Eccentricity~\cite{lin2024generating}  & 0.809 & 0.645 & 0.703 & 0.450 & 0.695 & 0.415 & 0.817 & 0.632 & 0.754 & 0.455 & 0.671 & 0.421 \\
GoogleSearchCheck ~\cite{chern2023factool}     & 0.672 & 0.470 & -- & -- & -- & -- & 0.779 & 0.673 & -- & -- & -- & -- \\
INSIDE~\cite{chen2024inside}  & 0.711 & 0.478 & 0.689 & 0.354 & 0.636 & 0.221 & -- & -- & -- & -- & -- & -- \\
Kernel Language Entropy\cite{nikitin2024kernel}   & 0.792 & 0.596 & 0.662 & 0.296 & 0.680 & 0.396 & 0.820 & 0.635 & 0.706 & 0.349 & 0.678 & 0.397 \\
SAPLMA \cite{azaria-mitchell-2023-internal}  & 0.850 & 0.726 & 0.815 & 0.642 & 0.651 & 0.347 & -- & -- & -- & -- & -- & -- \\
Semantic Entropy~\cite{kuhn2023semantic}  & 0.799 & 0.652 & 0.699 & 0.417 & 0.682 & 0.403 & 0.813 & 0.673 & 0.735 & 0.464 & 0.681 & 0.447 \\
Multi-LLM Collab~\cite{feng-etal-2024-dont}    & 0.632 & 0.350 & 0.689 & 0.320 & 0.681 & 0.347 & 0.778 & 0.565 & 0.933 & 0.879 & 0.671 & 0.399 \\
SAR~\cite{tokensar}    & 0.804 & 0.679 & 0.768 & 0.590 & 0.674 & 0.389 & 0.835 & 0.724 & 0.764 & 0.512 & 0.671 & 0.433 \\
Verbalized Confidence~\cite{tian-etal-2023-just}  & 0.759 & 0.547 & 0.579 & 0.234 & 0.698 & 0.460 & 0.836 & 0.740 & 0.652 & 0.369 & 0.717 & 0.514 \\
Directional Entailment Graph~\cite{da2024llmuncertaintyquantificationdirectional}
                        & 0.745 & 0.513 & 0.731 & 0.501 & 0.659 & 0.347 & 0.778 & 0.532 & 0.736 & 0.439 & 0.658 & 0.380 \\
\bottomrule
\end{tabular}
}
\caption{AUROC and PRR performance of UQ methods on TriviaQA, GSM8K, and FactScore-Bio across two models: LLaMA-3 8B and GPT-4o-mini.}
\label{table:results}
\end{table*}

\subsection{Calibration}
Calibration plays an important role in evaluating UQ performance by ensuring the scores produced by different approaches are comparable and interpretable by matching the range of the scores.
While some methods naturally output values within $[0,1]$, others may produce unbounded or differently scaled scores, which makes direct comparison across methods difficult. 
For example, some methods output values between 0 and 1, while others produce unbounded negative scores (e.g., in the range $(-\infty, 0]$). As a result, directly comparing raw scores can be challenging.

Calibration refers to the process of mapping raw scores into a normalized interval, typically $[0,1]$.
Several calibration techniques have been suggested.
Simple unsupervised approaches, such as min–max normalization, adjust scores based on their observed range, while more advanced supervised methods, such as isotonic regression~\cite{Han2017IsotonicRI}, learn mappings that better align predicted scores with correctness labels. 
Another widely used supervised method is sigmoid (Platt) calibration~\cite{guo2017calibrationmodernneuralnetworks}, which fits a logistic regression model to map raw scores into calibrated probabilities.

\section{Experiments}
\label{sec:Exp}

We evaluate the performance of a subset of available UQ methods through a series of experiments. In this section, we describe our experimental setup and discuss the results. We compare several widely used UQ approaches by using \texttt{TruthTorchLM} library \cite{yaldiz2025truthtorchlmcomprehensivelibrarypredicting}.

\subsection{Experimental Setup}
\noindent \textbf{Datasets.}\quad
Our primary evaluation focuses on short-form QA, a standard setting for assessing UQ performance. 
We use 1,000 samples from TriviaQA~\cite{joshi2017triviaqa} and GSM8K~\cite{cobbe2021gsm8k} for open-ended and mathematical reasoning questions, respectively. For long-form evaluation, we use FactScore-Bio~\cite{min-etal-2023-factscore}, which targets biographical questions with multi-fact generations.

\noindent \textbf{Models.}\quad
We conduct evaluations using both open- and closed-weight language models. Specifically, we use LLaMA-3-8B~\cite{llama3modelcard}, an open-source model that enables full access to internal states, and GPT-4o-mini~\cite{openai2023gpt4}, a closed-weight API model. We note that white-box UQ methods are not applicable to GPT-4o-mini.

\subsection{Discussion}

The results are summarized in Table~\ref{table:results}. 
Since each method entails different trade-offs including computational overhead, model access level, and supervision requirements, their performance varies accordingly. 
In short-form QA tasks (TriviaQA and GSM8K), LARS and SAPLMA achieve the highest performance, except on GSM8K with GPT-4o-mini, which is expected given that both are trained on labeled data. 
Among self-supervised methods, SAR performs the best on both TriviaQA and GSM8K for the LLaMA-3-8B model. For GPT-4o-mini, Verbalized Confidence achieves the best results on TriviaQA, while Multi-LLM Collab leads on GSM8K. 

FactScore-Bio evaluates long-form generation, which typically involves multiple factual claims and thus presents a more challenging setting for UQ. On this task, performance generally drops across methods compared to short-form QA. Verbalized confidence achieves the best results on both models. Eccentiricity and Semantic Entropy performs next best as sampling based methods, with Semantic Entropy showing stronger results for GPT-4o-mini. 
\section{Limitations and Future Directions of UQ}
\label{sec:Discussion}

In this section, we first examine the key limitations of existing LLM UQ methods, highlighting their current challenges and shortcomings. 
We then discuss possible lines for extending these approaches and outline promising directions for future research.

\subsection{Limitations of LLM Uncertainty Quantification}
Although existing UQ methods for LLMs have demonstrated effectiveness, they still face several critical limitations that hinder their practicality and applicability. 

\noindent \textbf{Knowledge Boundary of LLMs.}\quad
A central limitation stems from the inherent knowledge boundaries of LLMs~\cite{Huang_2025}. Many facts are inherently \emph{temporal}, meaning they evolve over time, whereas UQ scores are typically static with respect to a fixed model snapshot. This mismatch implies that UQ cannot always serve as a reliable tool for hallucination detection. For example, a query such as \emph{``How many goals has Messi scored this year?''} will naturally change as the year progresses. However, unless the LLM has access to updated external knowledge, its response will remain fixed based on its training data. In such cases, the model’s generation may become factually outdated (i.e., a hallucination), yet the associated uncertainty estimate will not reflect this temporal drift.

\noindent \textbf{Non-Interpretability of UQ Scores.}\quad
As discussed throughout the paper, there exist many algorithmically distinct UQ methods for LLMs. However, most of these methods do not produce scores within the intuitive range $[0,1]$, where 0 corresponds to 100\% wrong and 1 corresponds to 100\% correct. Instead, their raw outputs often lie on arbitrary scales, making them difficult to interpret directly as probabilities of correctness. While normalization or calibration techniques can partially address this issue, they typically require supervised or unlabeled reference data. Consequently, the interpretability of UQ scores remains a fundamental limitation across most methods.

\noindent \textbf{Difficulty of Training Multiple Models.}\quad
A principled way to quantify epistemic uncertainty is to train an ensemble of models under varying hyperparameters and training conditions, then measure the variability of their predictions for a given query. This approach is common in traditional classification tasks, where multiple models can be trained relatively cheaply. However, for LLMs, training even a single model is prohibitively expensive due to the immense computational and resource requirements. As a result, researchers are typically constrained to work with a single pre-trained model, limiting the feasibility of theoretically grounded ensemble-based approaches to UQ in the LLM setting.

\subsection{Future Directions of LLM Uncertainty Quantification}
While we have highlighted the current limitations of UQ, these challenges also reveal further opportunities for advancement. 
Existing approaches have demonstrated both effectiveness and shortcomings across diverse tasks, showing that no single method can fully address the wide spectrum of uncertainty in generative models. 
Consequently, there remains ample room for new lines of research that can broaden the scope, deepen the theoretical understanding, and enhance the practical reliability of UQ.

\noindent \textbf{Theoretically Grounded UQ Methods.}\quad
A key shortcoming of current UQ approaches for LLMs is that they largely rely on heuristics, often without clear connections to established uncertainty quantification theory or explicit distinctions between data (aleatoric) and model (epistemic) uncertainty. Unlike small-scale classification models, LLMs operate in a generative setting and introduce substantial practical challenges, such as the high cost of training and inference. As a result, many theoretically grounded approaches developed for smaller models cannot be directly applied. This highlights the need for new methods that both adhere to theoretical principles and remain feasible in practice, thereby enabling more rigorous and reliable science.

\noindent \textbf{UQ in Long-form QA.}\quad
Most existing UQ methods focus on short-form, open-ended QA. For example, questions such as ``What is the capital of France?'' can be answered concisely with ``Paris'' or ``The capital of France is Paris.''
However, in real-world applications, users often pose questions that demand more elaborate, long-form responses.
For instance, questions like ``Who is George Orwell?'' often require multiple claims, some of which are correct while others may be hallucinated, which clearly differs from short-form QA, where responses usually consist of a single claim.
Due to this critical difference, assigning a single uncertainty score to an entire long-form response is both impractical and undesirable, as it fails to reflect the correctness of individual claims within the text. 
Therefore, it is necessary to decompose the whole response into claims and assign UQ scores to each claim. 
Some strategies have been proposed to enable long-form UQ by modifying the query and response so that they align with claim-level evaluation~\cite{bakman-etal-2025-reconsidering}. These strategies include decomposing a long-form generation into individual claims, then applying UQ methods to these claims one by one. Although these strategies can be applied on top of existing UQ methods, their performance still lags behind short-form QA, motivating further research.

\noindent \textbf{Impact of Decoding Strategies on UQ.}\quad
Another underexplored direction lies in understanding how decoding strategies affect estimating uncertainty.
Decoding techniques such as temperature scaling and top-$k$ sampling directly reshape the probability distribution, thereby influencing UQ methods that directly rely on token probability.
Decoding also impacts output-consistency-based methods. 
For example, using low temperature narrows down the sampling space, and the model is more likely to generate consistent responses across multiple samples.
Nevertheless, such consistency does not necessarily imply correctness, since it stems from the decoding process, rather than from the model’s reliable knowledge of the correct answer.
These observations suggest that decoding choices should be explicitly accounted for in UQ research.

\noindent \textbf{Alternative Decompositions of LLM Uncertainty.} \quad
While much of the literature continues to adopt the traditional decomposition into epistemic and aleatoric uncertainty, this framework does not fully capture the nuances of interactive LLM systems. 
Thus, it raises a broader debate on how uncertainty in LLMs should be conceptualized.
A recent work~\cite{kirchhof2025position} argues that the binary categorization of uncertainty into aleatoric and epistemic is insufficient to fully capture the challenges posed by LLMs.
Instead, they introduce additional types of uncertainty that better reflect the interactive and open-ended nature of these systems: \textit{task-underspecification uncertainty}, which arises when a prompt fails to clearly specify the task, and \textit{context-underspecification uncertainty}, which occurs when the input query omits essential information required for a correct answer. Since these proposals remain at a conceptual level without explicit formalization, developing new frameworks and decompositions for UQ in LLMs represents an important direction for future research.

\noindent \textbf{Novel Applications of UQ.} \quad
The primary purpose of UQ is to assess the reliability of LLM generations, but it can be extended to various applications for more reliable and trustworthy generation.
One promising application is \textit{uncertainty-aware adaptive guidance} for reasoning tasks, where the model can backtrack from an uncertain reasoning step to a more reliable point and incorporate additional demonstrations to correct the reasoning process~\cite{yin-etal-2024-reasoning}. 
Similarly, UQ can be leveraged in language-agent settings: when the model detects high uncertainty, it can explicitly acknowledge this, defer an answer, or call external tools to obtain more reliable information~\cite{han-etal-2024-towards}. 
Another line of application involves directly intervening in the decoding process by adjusting the sampling strategy, such as generating multiple candidate continuations and selecting the one with the highest confidence~\cite{chuang-etal-2024-lookback}. Lastly, UQ methods can be used as reward models, so they can even be used in reinforcement learning–based fine-tuning of LLMs.
\section{Conclusion}
In this survey, we present a comprehensive examination of uncertainty quantification (UQ) in large language models (LLMs) with a particular focus on hallucination detection. 
We outline the foundations of UQ in traditional machine learning classification and then discuss how these notions are adapted to the context of LLMs, connecting the role of UQ to hallucination detection.
We then systematically categorize a wide range of existing UQ methods along multiple dimensions---including conceptual approach, sampling requirements, model accessibility, and training reliance---and provide concise explanations for each category. 
We further describe the key concepts for UQ evaluation, and provide empirical results for representative approaches.
Beyond quantitative evaluation, we also examine the limitations that current UQ methods face and outline future research directions aimed at addressing these challenges, with the goal of advancing more robust and generalizable UQ approaches.
Overall, this survey consolidates the current landscape of UQ for hallucination detection, offering readers both a clear and structured understanding of existing techniques and a roadmap for future progress. 
We hope that this work serves as a useful reference for researchers and practitioners striving to build more reliable and trustworthy LLM systems.

\bibliographystyle{plainnat}
\bibliography{references}

\end{document}